\documentclass[sigconf]{acmart}
\AtBeginDocument{%
  }

\copyrightyear{2026}
\acmYear{2026}
\setcopyright{cc}
\setcctype{by}
\acmConference[MM '26]{Proceedings of the 34th ACM International Conference on Multimedia}{November 10--14, 2026}{Rio de Janeiro, Brazil}
\acmBooktitle{Proceedings of the 34th ACM International Conference on Multimedia (MM '26), November 10--14, 2026, Rio de Janeiro, Brazil}
\acmDOI{10.1145/3767308.3836122}
\acmISBN{979-8-4007-2213-4/2026/11}

\usepackage{algorithm}
\usepackage{algpseudocode}
\usepackage{amsmath}
\usepackage{booktabs}
\usepackage{multirow}
\usepackage{pifont}
\usepackage{ulem}
\usepackage{cuted}
\begin{document}

\title{Semantic-Spatial Discriminability Enhancement for Generalized Visual Grounding}

\author{Kaiyan Lei}
\orcid{0009-0005-4964-3570}
\affiliation{%
  \institution{State Key Laboratory of Multimodal Artificial Intelligence Systems, Institute of Automation, \\Chinese Academy of Sciences}
  \city{Beijing}
  \country{China}
}
\email{leikaiyan2025@ia.ac.cn}

\author{Xu-Yao Zhang}
\authornote{Corresponding author}
\affiliation{%
  \institution{State Key Laboratory of Multimodal Artificial Intelligence Systems, Institute of Automation, \\Chinese Academy of Sciences}
  \city{Beijing}
  \country{China}
}
\email{xyz@nlpr.ia.ac.cn}

\renewcommand{\shortauthors}{Kaiyan Lei and Xu-Yao Zhang}

\begin{abstract}
  Generalized Visual Grounding (GVG) task aims to localize targets in an image based on referring expressions, extends the classical visual grounding paradigm by integrating multi-target and non-target scenarios. 
Previous methods typically rely on global semantic matching or coarse-grained region interactions for localization, where the discriminative cues are primarily derived from sentence-level semantics or regional context. In complex multi-target scenarios, such approaches tend to confuse visually similar targets, making it difficult to establish stable instance-level decision boundaries.
To address these limitations, this paper proposes a novel Semantic–Spatial Discriminability Enhancement (SSDE) framework for generalized visual grounding, which aims to enhance the discriminative ability on fine-grained semantics and spatial localization, improving both cross-modal understanding and instance-level grounding.
Specifically, to enhance the semantic discriminability of query representations at the fine-grained level,
we propose a Semantic Discriminability Enhancement (SeDE) module, which leverages spatially guided cross-attention to disentangle fine-grained target-relevant visual attributes and integrates them with the textual subject semantics.
Furthermore, to strengthen the spatial discriminability of the referred targets, 
we introduce a Spatial Discriminability Enhancement (SpDE) module, which models an instance center density map to characterize the spatial distribution of targets, and explicitly constructs instance separation structures in the spatial domain by employing them as an auxiliary supervision signal.
Extensive experiments show that SSDE achieves superior performance on ten datasets across both classic and generalized visual grounding tasks. Code will be available at \url{https://github.com/Letitialky/GVG-SSDE}.

\end{abstract}




\maketitle

\section{Introduction}

\begin{figure}[t]
  \centering
  \includegraphics[width=\linewidth]{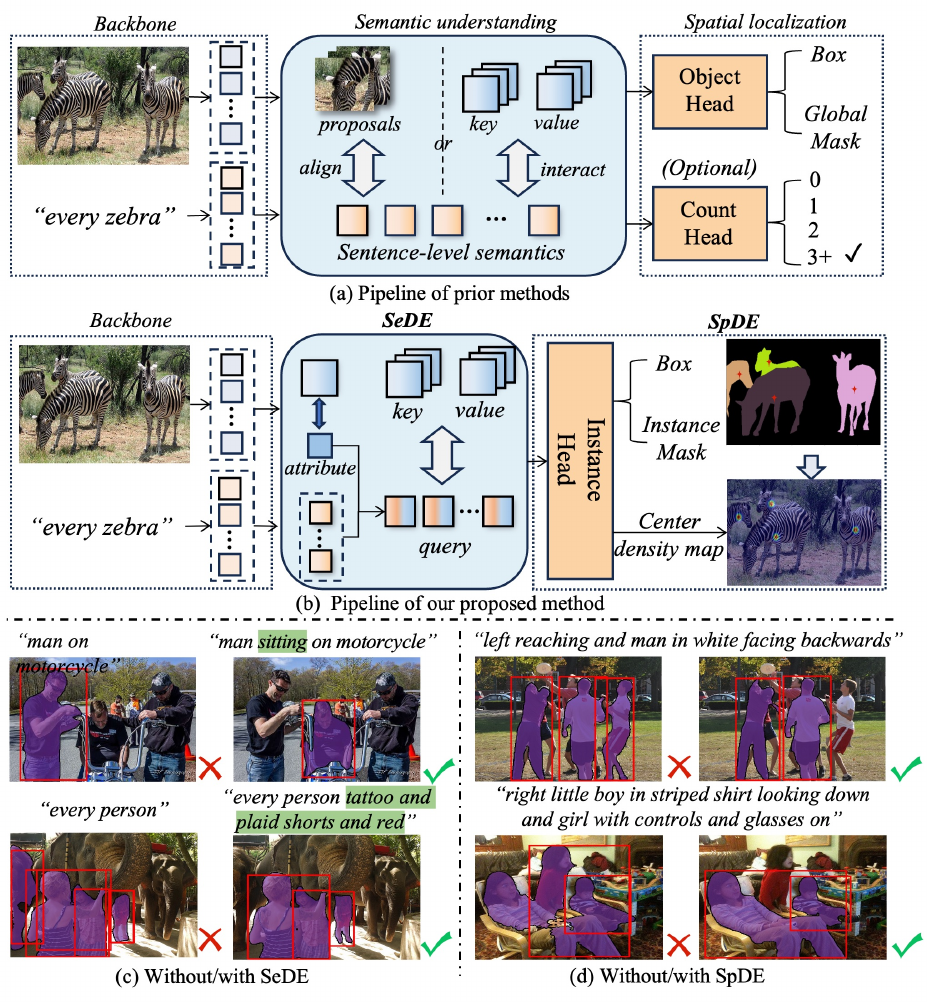}
  \caption{Motivation visualization. (a) Pipeline of prior methods, which rely on sentence-level semantics, predict global masks and optionally use a count head. (b) Pipeline of SSDE, which proposes SeDE to enhance semantic discriminability and proposes SpDE to enhance spatial discriminability.
(c) The two columns show the results without and with SeDE. 
(d) The two columns show the results without and with SpDE.}
  \label{fig_1}
\end{figure}

Visual Grounding (VG) aims to localize all targets in an image that satisfy the semantics specified in a referring expression. In Classic Visual Grounding (CVG), each expression corresponds to one single target, comprising two subtasks: Referring Expression Comprehension (REC) \cite{yu2018mattnet,shi2023dynamic,kamath2021mdetr,dai2024simvg}, which requires predicting target bounding box, and Referring Expression Segmentation (RES) \cite{shang2024prompt,huang2020referring,liu2023polyformer,cheng2024parallel}, which requires pixel-level segmentation. To better reflect real-world scenarios, Generalized Visual Grounding (GVG) \cite{liu2023gres,hu2023beyond,wu2024toward} further extends CVG to handle multi-target and non-target cases through Generalized Referring Expression Segmentation (GRES) and Generalized Referring Expression Comprehension (GREC). 
This setting introduces additional challenges, including open-ended quantities, multi-target semantic relations, and intricate scene structures. It is required to establish stable decision boundaries for cross-modal matching between potential targets and the referring expression within an open and uncertain search space, thereby imposing higher demands on model's discriminative capability.

As illustrated in Figure \ref{fig_1}, we analyze discriminability from both semantic and spatial perspectives. From a semantic perspective, expressions are typically concise and label-like, such as collective references (e.g., “everyone”) and modifier-based expressions (e.g., “man on motorcycle”). The former requires identifying multiple targets sharing a common category, whereas the latter demands precise discrimination among visually similar candidates. Existing methods typically rely on sentence-level semantics, modeling the complex text-image correspondence either through alignment with region proposals \cite{yu2018mattnet,liu2019improving,yao2024visual,hong2019learning,dai2025propvg} or interaction with visual embeddings \cite{liu2023polyformer,su2023language,deng2021transvg,liu2023caris,liu2023gres,dai2024simvg,zheng2025look}. However, real-world scenes often contain numerous distractor regions. Relying solely on coarse-grained textual semantics makes it difficult to distinguish “the best-matching targets” from “partially relevant candidates,” leading to mislocalization or over-response. Therefore, one of our research objectives is \textbf{\textit{(1) how to enlarge the discriminative margins among similar targets in the semantic space to enhance semantic discriminability.}}
From a spatial perspective, flexible and diverse expressions may refer to an arbitrary number of targets from varying viewpoints (e.g., “all people” or "the whole group"), requiring to not only determine targets presence but also identify the individual spatial locations. Notably, the target number is an implicit property of spatial structure, and its uncertainty directly affects localization. Prior methods \cite{liu2023gres,liu2023caris,luo2025cohd,yu2025latent,wang2025hierarchical} typically predict a global mask, which essentially projects multiple discrete targets into a unified region. Some works \cite{luo2025cohd,wang2025hierarchical} further discretize the object counts into predefined categories and use an count head to explicitly predict the global count via classification or regression. However, multi-target scenarios involve substantial inter-instance competition and overlap, where the global mask fail to provide stable instance-level decision boundaries. And subtle differences in object count are seldom effectively propagated back to guide spatial localization.
Therefore, our second objective is
\textbf{\textit{(2) how to characterize instance-level spatial separability and structural independence among multiple targets in the spatial domain, thereby improving spatial discriminability}.}

Regarding the enhancement of semantic discriminability, when a single expression encompasses multiple highly similar targets within the same category (e.g., “everyone”), distinguishing among them often depends on fine-grained visual cues such as color, shape, or local texture. As illustrated in Figure \ref{fig_1}(c), in the presence of similar candidate targets, a description like “man on motorcycle” tends to incorrectly highlight the salient instance, whereas augmenting the expression with an attribute “sitting” yields accurate localization. Moreover, expression like “every person” yields redundant predictions, whereas incorporating attribute descriptors such as “tattoo,” “plaid shorts,” or “red” leads to correct and complete localization. These observations demonstrate that visual attributes, under fine-grained relational constraints, effectively mitigate semantic ambiguity and facilitate consistent and comprehensive identification of target sets in multi-target scenarios.

Regarding the enhancement of spatial discriminability, InstanceVG \cite{dai2025improving} introduces instance-level supervision to impose fine-grained spatial constraints. Although it improves accuracy, such pixel-wise supervision is spatially dense signals, making it more suitable for pixel classification while still insufficient for capturing stable multi-instance spatial structures. As illustrated in Figure \ref{fig_1}(d), the 1st column shows results under instance-level supervision alone. When localizing “left reaching and man in white facing backwards”, it has redundant predictions due to its lack of spatial structural awareness. In another case, it is difficult to define accurate instance boundaries under severe instance adhesion. In essence, count awareness corresponds to identifying object centers \cite{kang2024vlcounter,jiang2023clip}. By explicitly modeling instance centers, as shown in the 2nd column, it is able to maintain instance-level structural separability and accurate localization, even under ambiguous boundaries or partial occlusions.

Based on these analysis, we propose a novel \textbf{Semantic–Spatial Discriminability Enhancement (SSDE)} framework for GVG task, aiming to jointly improve semantic and spatial discriminability, thereby advancing cross-modal understanding and instance-level localization.
Specifically, we introduce a \textbf{Semantic Discriminability Enhancement (SeDE)} module, which leverages spatially guided cross-attention to extract target-relevant visual attribute representations and integrate them with the textual semantics. Its primary purposes are:
(1) to enhance the discriminative capacity of query representations by explicitly incorporating fine-grained visual attributes;
(2) to enlarge inter-instance discriminability within the cross-modal embedding space;
(3) to mitigate missed detections and ambiguity arising from insufficient visual cues.
In addition, we propose a \textbf{Spatial Discriminability Enhancement (SpDE)} module, which couples spatial and count awareness by explicitly modeling the “text-to-instance” mapping through instance center responses. Beyond the region-level constraints provided by instance-level supervision, it further imposes instance-level structural constraints as an auxiliary objective. Its main purposes are:
(1) to ensure spatial separability and independence among multiple targets;
(2) to guide feature learning via predicted spatial distributions as an implicit prior;
(3) to alleviate instance adhesion and spatial ambiguity in complex scenes.
We conduct extensive experiments on popular CVG and GVG benchmarks, including RefCOCO/+/g \cite{kazemzadeh2014referitgame,mao2016generation}, gRefCOCO \cite{liu2023gres}, Ref-ZOM \cite{hu2023beyond}, and R-RefCOCO \cite{wu2024toward}. The results demonstrate that our method significantly outperforms prior approaches across all benchmarks, validating its effectiveness in handling complex GVG scenarios.

Overall, our contributions are summarized as follows.

\begin{itemize}
\item We propose a novel semantic–spatial discriminability enhancement framework for GVG, which improves cross-modal understanding and instance-level localization by jointly strengthening semantic and spatial discrimination.
\item We propose a semantic discriminability enhancement module that explicitly incorporates fine-grained visual attributes via spatially guided cross-modal interactions, thereby enhancing inter-instance discriminability.
\item We propose a spatial discriminability enhancement module, which models instance center as an auxiliary objective to impose instance-level structural constraints, ensuring the separability and independence of multiple instances.
\item Extensive experiments demonstrate that the proposed method achieves state-of-the-art performance across ten various CVG and GVG benchmarks, delivering significant improvements over the existing methods.
\end{itemize}

\begin{figure*}
\centering
\includegraphics[width=7.1in]{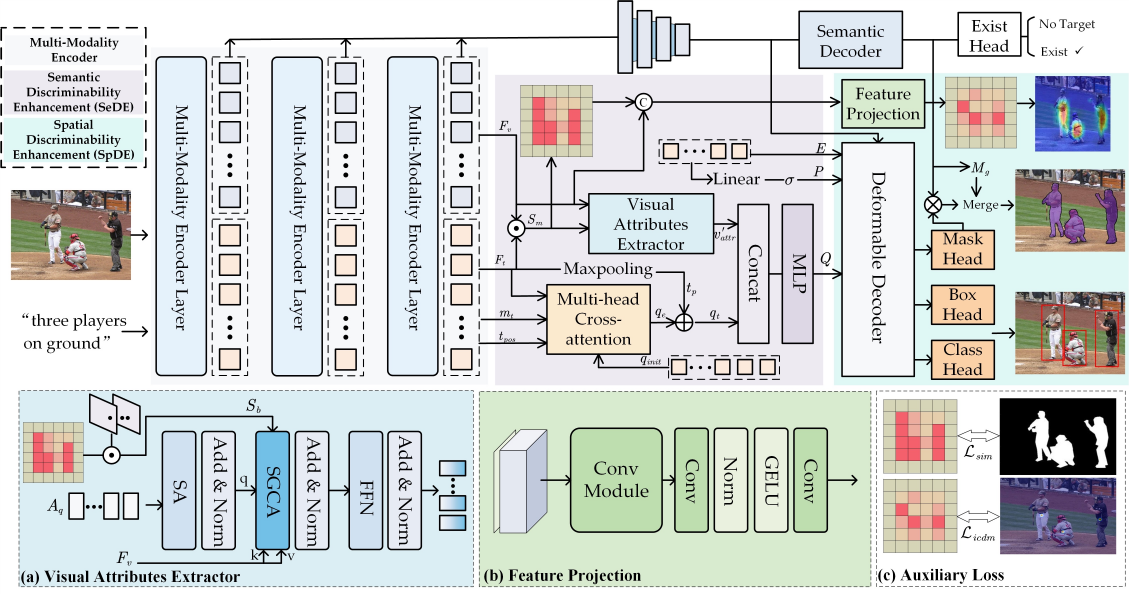}
\caption{ Overall architecture of SSDE. The framework consists of a multi-modality encoder backbone, a Semantic Discriminability Enhancement (SeDE) module, a Spatial Discriminability Enhancement (SpDE) module.
(a) Illustration of the visual attribute extraction process.
(b) Convolution-based feature mapping process in the instance center prediction.
(c) Supervision applied to the similarity map and the instance center density map during training.}
\label{fig_2}
\end{figure*}

\section{Related Work}

\subsection{Classic Visual Grounding}

The CVG task addresses the localization of a single target. Early approaches achieve text–object matching by generating candidate region proposals \cite{yu2018mattnet,hong2019learning,chen2021ref,yu2018rethinking,liu2019improving,zhang2018grounding,yao2024visual} or by performing matching over dense anchor-based detections \cite{yang2020improving,zhou2021real,yang2019fast,luo2020multi}. Subsequently, Transformer-based methods \cite{kim2022restr,hu2025remerec,su2023language,xiao2024hivg,dai2024simvg,shang2024prompt,zhu2022seqtr,liu2023polyformer,liu2024mapper} have significantly improved multi-modal understanding by modeling cross-modal relationships between visual and textual representations. These methods typically adopt an encoder–decoder architecture. Specifically, some methods \cite{deng2021transvg,deng2023transvg++,chng2024mask,kamath2021mdetr,luo2025cohd,wang2025hierarchical} first encode the features of each modality independently using visual and text backbones, (e.g., ViT\cite{dosovitskiy2020image}, Swin Transformer \cite{liu2021swin}, BERT\cite{devlin2019bert}, RoBERTa \cite{liu2019roberta} etc.) and then perform multimodal fusion within the encoder, while others \cite{su2023language,yang2022lavt,ye2022shifting} integrate the multimodal fusion process directly into the visual backbone, enabling earlier cross-modal interaction. However, since the modality features are extracted from pretrained models trained on unrelated tasks like classification or regression, cross-modal understanding in these frameworks still relies heavily on limited downstream training data. With the rapid development of vision–language pretraining \cite{radford2021learning,oquab2023dinov2}, several methods \cite{xiao2023clip,xiao2024hivg,wang2024referencing,shang2024prompt,shi2023dynamic,shi2025swimvg} have emerged that adopt pretrained CLIP encoders to improve performance through fine-tuning or adaptation strategies. Nevertheless, approaches based on vision–language pretrained models place greater emphasis on modal alignment, which motivates recent researches \cite{dai2024simvg,xiao2024oneref,yu2025latent,dai2025improving,dai2025multi} toward unified multi-modality encoders that simultaneously encode modality features while enhancing cross-modal semantic interaction. Furthermore, according to the form of predictions, CVG can be divided into two sub-tasks of REC and RES. Because solving these similar yet distinct tasks typically requires task-specific and complex model designs, recent studies \cite{luo2020multi,li2021referring,dai2025multi,chen2024efficient,zhao2024rethinking,dai2025improving} attempt to develop multi-task VG frameworks, aiming to exploit complementary supervision signals across tasks to jointly address object detection and segmentation.

\subsection{Generalized Visual Grounding}

In GVG task, expressions may refer to zero, single, or multiple targets. DMMI \cite{hu2023beyond} introduces a segmentation benchmark that extends single-target to multi-target settings. RefSegformer \cite{wu2024toward} further considers both positive and negative expressions, improving robustness in distinguishing non-target regions. ReLA \cite{liu2023gres} formalizes the GRES task, which fully generalizes the multi-/single-/non- target cases. It partitions the input image into regions and performs soft aggregation over region features to establish long-range region–region dependencies in multi-target scenarios. RaAM-VG \cite{ouyang2025region} introduces learnable region-aware anchors to guide attention toward targets, alleviating cross-modal redundancy that may arise when direct visual-language interaction causes the model to prioritize the most salient objects. Some other works focus on improving semantic understanding of targets. MattNet \cite{yu2018mattnet} performs attribute parsing on the expressions to enhance targets understanding. LatentVG \cite{yu2025latent} generates latent expressions based on visual details to enrich the target semantics of a single-text expression. In addition, several studies focus on improving target counting. COHD \cite{luo2025cohd} enables the model to learn counting abilities for specific categories under multi-/single-/non-target conditions, thereby reducing ambiguity caused by the inherent differences between single and multiple targets. Similarly, HieA2G \cite{wang2025hierarchical} introduces an explicit multi-class classifier to determine the number of output targets for each image–text pair. In terms of model architecture, \cite{dai2024simvg,dai2025propvg,dai2025improving,yu2025latent} adopt a unified multi-modality encoder backbone and achieve superior performance on the GVG task. SimVG \cite{dai2024simvg} employs a weight-distillation-based multi-branch synchronous learning strategy to enhance the representation capability of its lightweight architecture. PropVG \cite{dai2025propvg} improves proposal-based frameworks to strengthen the target discriminability. InstanceVG \cite{dai2025improving} introduces instance-level spatial fine-grained perception and incorporates instance segmentation supervision, thereby enhancing the spatial understanding.

\section{Methods}

\subsection{Overview}

We develop SSDE based on the instance segmentation framework \cite{cheng2022masked,li2023mask}, as illustrated in Figure \ref{fig_2}. We adopt the multi-modality encoder BEiT-3 \cite{wang2023image} to jointly process the input image $I \in \mathbb{R}^{H \times W \times 3}$ and the referring expression $T$, where H and W is the height and width, producing aligned visual and textual features through unified vision-language encoding. The SeDE generates queries enriched with fine-grained visual attribute representations, enhancing semantic discrimination. The SpDE predicts an instance center density map, which is jointly optimized with instance-level supervision as an auxiliary loss to enforce structural constraints.  

Specifically, the encoder is organized into three stages, yielding hierarchical visual features $\{F_v^1, F_v^2, F_v^3\}$. All visual and textual features are projected into a d-dimensional space using linear layers, resulting in $\{F_v', F_v'', F_v\} \in \mathbb{R}^{hw \times d}$ and $F_t \in \mathbb{R}^{L \times d}$, where $h=\frac{H}{p}$, $w=\frac{W}{p}$, $p$ is patch size, and $L$ is the text length.
We further construct multi-scale visual features using a feature pyramid network (FPN) \cite{li2022exploring}. Specifically, $F_v'$ and $F_v''$ are upsampled via transposed convolution to obtain $\widehat{F_v'} \in \mathbb{R}^{4h \times 4w \times d}$ and $\widehat{F_v''} \in \mathbb{R}^{2h \times 2w \times d}$, respectively, while $F_v$ is downsampled via max-pooling to produce $\widehat{F_v} \in \mathbb{R}^{\frac{h}{2} \times \frac{w}{2} \times d}$. The final multi-scale feature set is defined as $\mathcal{F} = \{\widehat{F_v'}, \widehat{F_v''}, F_v, \widehat{F_v}\}$.
Finally, a lightweight semantic decoder composed of convolutional layers aggregates multi-scale features into global semantic features $F_g$, which are further transformed into a global segmentation map $M_g$. The instance segmentation map is obtained by multiplying $F_g$ with features of Mask Head, and then merging with $M_g$ to produce the final segmentation prediction.
For detection, Box and Class Head produce the final bounding box predictions, while an additional Exist Head, composed of MLPs, operates on $F_g$ to perform binary classification for target presence. The decoder is implemented as a deformable decoder \cite{zhu2020deformable}.

\subsection{Semantic discriminability enhancement}
Textual expressions typically provide only coarse-grained subject information. To obtain precise discrimination cues in scenarios with densely distributed similar candidates, we augment textual semantics with target-relevant visual attributes. We leverage a spatially guided attention mechanism to extract fine-grained visual attributes and fuse them with subject-level semantic representations to generate more discriminative query representations, thereby enhancing the semantic discriminability between instances.

\textbf{Textual subject semantic extraction.}
To extract the subject-level semantics from the referring expression, we introduce $N_q$ learnable query embeddings 
$q_{init} \in \mathbb{R}^{N_q \times d}$. These embeddings interact with the textual features $F_t \in \mathbb{R}^{L \times d}$ through cross-attention mechanism to aggregate token-level semantic information, yielding text-driven queries $ q_t $:

\begin{equation}
q_t = \mathrm{CrossAttn}(q_{init}, F_t).
\end{equation}
In addition, to further incorporate global context, we apply max pooling operation over $F_t$ to highlight salient semantic components, producing $ t_p $.
The final subject-aware textual representation is $ q_e = q_t + t_p $, which jointly captures both local token-level semantics and global contextual information.

\textbf{Visual attribute extraction.}
We exploit the image-text similarity map as a spatial prior and employ a spatially-guided attention mechanism to focus on potential target regions, thereby further extracting fine-grained visual attributes. 
Specifically, given the visual features $F_v \in \mathbb{R}^{hw \times d}$ and textual features $F_t$, we first compute the cross-modal similarity $sim = F_v F_t^{\top}$.
Then, average pooling is applied along the textual dimension followed by a $1\times1$ convolution to obtain a spatial similarity map:

\begin{equation}
S_m = \mathrm{Conv}_{1\times1}\big(\mathrm{MeanPool}(sim \odot m_t)\big),
\end{equation}
where $m_t \in \{0,1\}^{L}$ denotes the text validity mask. The resulting map $S_m \in \mathbb{R}^{h\times w}$ represents the spatial semantic correspondence between image and text. During training, $S_m$ is supervised by the ground-truth mask to ensure accurate indication of target regions.

To further incorporate this spatial prior into the attention mechanism, we construct a bias term $S_b$ by transforming $S_m$ into a spatial attention distribution:

\begin{equation}
S_{b} = \mathrm{Softmax}(m \cdot S_m),
\end{equation}
where $m$ is a learnable scaling factor that adaptively adjust the strength of the bias term. This operation explicitly guides attention toward semantically relevant regions.

Based on this design, we introduce Spatially-Guided Cross Attention (SGCA), which extends the standard cross-attention\cite{vaswani2017attention} by injecting the spatial bias into the attention logits, which enhances aggregation over target-relevant regions.

\begin{equation}
\text{Attn} =
\mathrm{Softmax}\left(
\frac{QK^{\top}}{\sqrt{d}} + S_{b}
\right).
\end{equation}
Finally, we introduce $N_q$ learnable attribute tokens 
$A_q \in \mathbb{R}^{N_q \times d}$, which interact with $ F_v $ via SGCA to extract fine-grained attributes such as color, texture, and pose from target-related regions:

\begin{equation}
v_{attr} = \mathrm{SGCA}(A_q, F_v).
\end{equation}

\begin{algorithm}[t]
\caption{Instance Center Density Map Construction}
\label{alg:density_map}
\begin{algorithmic}[1]
\Require Instance masks $\mathcal{M}=\{M_1,M_2,\dots,M_K\}$, image size $(H,W)$, Gaussian parameter $\sigma$
\Ensure Density map $D$

\State Initialize density map: $D \leftarrow \mathbf{0}_{H\times W}$

\For{each instance mask $M_k \in \mathcal{M}$}

    \State Perform connected component analysis: \\
    $\{C_{k,1},C_{k,2},\dots,C_{k,n_k}\}$

    \If{$n_k = 0$}
        \State \textbf{continue}
    \EndIf

    \State Compute the area of each component:
    $A_{k,i} = \sum C_{k,i}$

    \State Select the largest component:
    $i^* = \arg\max_i A_{k,i}$

    \State Extract component pixels:
    $C_k^* = C_{k,i^*}$

    \State Compute center coordinate of $C_k^*$:
    $c_k = (c_x^{k}, c_y^{k})$

    \State Generate Gaussian distribution:
    \[
    G(x,y) = \exp\left(
    -\frac{(x-c_x^{k})^2 + (y-c_x^{k})^2}{2\sigma^2}
    \right)
    \]

    \State Update density map:
    $D = \max(D, G)$

\EndFor

\State \Return $D$

\end{algorithmic}
\end{algorithm}

\textbf{Query generation.} 
Finally, we generate a more discriminative query representation, in which subject semantics determine the category while attribute details enable instance-level differentiation. Specifically, the attribute features $v_{attr}$ are first transformed through an MLP to enhance their expressive capacity, yielding $ v_{attr}^{'} = MLP(v_{attr}) $. The transformed features are then concatenated with
$q_e$, and fused through an MLP to produce the final attribute-enhanced queries $ Q=q_{i=1}^{N_q}, q\in R^{N_q\times d}$:

\begin{equation}
q = \mathrm{MLP}([q_e; v_{attr}^{'}]).
\end{equation}

We adopt a deformable decoder\cite{zhu2020deformable}, where sampling locations guide the attention mechanism. To enhance the representational flexibility of global learnable queries, we introduce a set of learnable positional embedding 
$E = \{e_i\}_{i=1}^{N_q}$, where $e_i \in \mathbb{R}^d$ shares the same dimensionality as the $q_i$ and is optimized adaptively via backpropagation during training.
The positional embeddings are further projected into a two-dimensional spatial domain through a linear transformation, followed by a sigmoid function to constrain the coordinates within $[0,1]$, yielding learnable reference points:
\begin{equation}
p_i = \sigma(W e_i),
\end{equation}
where $W$ denotes learnable parameters and $\sigma(\cdot)$ is the sigmoid function. The resulting reference point set is defined as:
\begin{equation}
P = \{p_i\}_{i=1}^{N_q}.
\end{equation}
These reference points serve as the initial sampling locations for deformable attention, guiding sparse and effective feature aggregation over multi-scale features. During decoding, $P$ is used as reference points, $E$ as the positional encodings, and $Q$ as the query features, which are jointly fed into the deformable decoder.

\subsection{Spatial discriminability enhancement}
To enhance spatial discriminability in dense scenes and mitigate spatial ambiguity, we explicitly model instance center responses to accurately represent instance-level spatial structures, and predict instance centers via a lightweight feature projection module. This process serves as an auxiliary loss that imposes instance-level structural constraints, thereby ensuring spatial separability across instances and improving instance-level spatial discrimination.

\begin{table*}[!htbp]
\caption{Comparison with state-of-the-art methods on the RefCOCO/+/g \cite{yu2016modeling} datasets for the RES task. In the “Backbone” column, a “/” indicates separate backbones for visual and textual modalities. “FT” indicates whether fine-tuning is performed on the target dataset, and “MT” denotes whether the model is trained in a multi-task setting. mIoU is used as the evaluation metric. The best results are highlighted in bold, and the second-best results are underlined.}
\centering
\resizebox{\linewidth}{!}{
    \begin{tabular}{*{13}{c}}
        \toprule
        \multirow{2}*{Models} & \multirow{2}*{Publication} & \multirow{2}*{Backbone} & \multirow{2}*{FT} & \multirow{2}*{MT} & \multicolumn{3}{c}{RefCOCO} & \multicolumn{3}{c}{RefCOCO+} & \multicolumn{2}{c}{RefCOCOg} \\ 
        \cmidrule(lr){6-8} \cmidrule(lr){9-11} \cmidrule(lr){12-13} 
        & & & & & val & testA & testB & val & testA & testB & val(u) & test(u) \\
      
        \midrule
        \multicolumn{13}{c}{MLLM Methods} \\
        \midrule
        LISA-L2-13B \cite{lai2024lisa} & CVPR24 & SAM-ViT-H & \ding{51} & \ding{55} & 76.30 & 78.70 & 72.40 & 66.20 & 71.00 & 59.30 & 70.10 & 71.10  \\
        
        GSVA-L2-13B \cite{xia2024gsva} & CVPR24 & SAM-ViT-H & \ding{51} & \ding{55} & 79.20 & 81.70 & 77.10 & 70.30 & 73.80 & 63.60 & 75.70 & 77.00  \\
        
        GLaMM-7B \cite{rasheed2024glamm} & CVPR24 & CLIP-ViT-H & \ding{51} & \ding{55} & 79.50 & 83.20 & 76.90 & 72.60 & 78.70 & 64.60 & 74.20 & 74.90  \\
        \midrule
        \multicolumn{13}{c}{Specialist Methods} \\
        \midrule
        MCN \cite{luo2020multi} & CVPR20 & DarkNet-53/GRU & \ding{55} & \ding{51} & 62.44 & 64.20 & 59.71 & 50.62 & 54.99 & 44.69 & 49.22 & 49.40 \\
        
        ReLA \cite{liu2023gres} & CVPR23 & Swin-B/BERT & \ding{55} & \ding{55} & 73.82 & 76.48 & 70.18 & 66.04 & 71.02 & 57.65 & 65.00 & 65.97 \\
        
        EEVG \cite{chen2024efficient} & ECCV24 & ViT-B/BERT & \ding{51} & \ding{51} & 79.49 & 80.87 & 77.39 & 71.86 & 76.67 & 66.31 & 73.56 & 73.47 \\
        
        PromptRIS \cite{shang2024prompt} & CVPR24 & SAM-CLIP-B/CLIP & \ding{55} & \ding{55} & 78.10 & 81.21 & 74.64 & 71.13 & 76.60 & 64.25 & 69.17 & 70.47 \\
       
        HieA2G \cite{wang2025hierarchical} & AAAI25 & Swin-B/RoBERTa & \ding{51} & \ding{51} & 75.10 & 77.60 & 71.10 & 66.50 & 71.40 & 58.90 & 65.30 & 66.60 \\

        COHD \cite{luo2025cohd} & ICCV25 & Swin-B/BERT & \ding{55} & \ding{55} & 78.11 & 80.39 & 75.20 & 72.03 & 76.37 & 65.45 & 70.83 & 72.11 \\

        RaAM-RVG \cite{ouyang2025region} & ICCV25 & ViT-B/BERT & \ding{55} & \ding{51} & 79.35 & 81.22 & 77.81 & 69.54 & 75.69 & 63.02 & 71.30 & 72.09  \\
        \midrule
        OneRef-B \cite{xiao2024oneref} & NerulPS25 & BEiT3-ViT-B & \ding{51} & \ding{51} & 79.83 & 81.86 & 76.99 & 74.68 & 77.90 & 69.58 & 74.06 & 74.92 \\
         
        Latent-VG \cite{yu2025latent} & ICCV25 & BEiT3-ViT-B & \ding{55} & \ding{51} & 81.01 & 82.26 & \uline{79.77} & \uline{76.92} & 79.48 & \uline{72.95} & \uline{76.10} & 76.51 \\

        PropVG \cite{dai2025propvg}& ICCV25 & BEiT3-ViT-B & \ding{55} & \ding{51} & 77.99 & 79.81 & 75.28 & 72.94 & 76.49 & 67.22 & 71.34 & 72.10\\ 
 
        InstanceVG \cite{dai2025improving} & TPAMI25 & BEiT3-ViT-B & \ding{55} & \ding{51} & \uline{81.36} & \uline{83.05} & 79.28 & 76.64 & \uline{79.51} & 71.56 & 75.89 & \uline{76.59} \\
        
        \midrule
        Ours & - & BEiT3-ViT-B & \ding{55} & \ding{51} & \textbf{81.82} & \textbf{83.34} & \textbf{79.83} & \textbf{77.44} & \textbf{80.51} & \textbf{74.00} & \textbf{77.08} & \textbf{77.80} \\
        \bottomrule
        
    \end{tabular}}
    \label{tab_3}
\end{table*}

\textbf{Instance center density map construction.} 
We construct an instance center density map based on instance segmentation annotations. Specifically, the density map is generated by placing a two-dimensional Gaussian kernel at the center of each instance, forming a continuous representation of object centers.
The determination of instance centers is critical, as it directly affects the encoded spatial structure and the training stability when it used as the supervision signal. Given an instance mask $M_k$, we first perform connected component analysis and define the instance center $c_k = \left(c_x^{k}, c_y^{k}\right)$ as the centroid of the largest connected component, and the coordinates are computed as:

\begin{equation}
c_x^{k} = \frac{1}{|C_k^*|} \sum_{(x,y) \in C_k^*} x, 
\quad
c_y^{k} = \frac{1}{|C_k^*|} \sum_{(x,y) \in C_k^*} y,
\end{equation}
where $C_k^*$ denotes the largest connected component of $M_k$. The rationale is that the largest component typically preserves the most complete and salient structure of the instance, thereby providing a clearer and more stable supervisory signal.

After determining the center locations, each discrete center is expanded into a continuous spatial density signal using a two-dimensional Gaussian kernel. For images containing multiple instances, the Gaussian responses corresponding to different centers are aggregated by a point-wise maximum operation:

\begin{equation}
D(x,y) = \max_{k=1,\dots,K} G_k(x,y),
\end{equation}
where $D(x,y)$ denotes the final instance center density map. The detailed construction procedure is summarized in Algorithm 1.
The resulting density map naturally encodes the counts information of multiple targets while providing explicit supervision of instance structures. In multi-instance scenarios, it simultaneously represents all instance centers, allowing subtle variations in target count to be reflected in the spatial localization optimization process.

\textbf{Instance center prediction.}
We introduce an instance-centered prediction module that enables the model to leverage spatial distribution predictions as an implicit prior for feature learning, while using the density map $D$ as a stable and semantically explicit supervisory signal. Specifically, the high-response regions in the similarity map $S_m$ typically correspond to the entire target regions. To further focus on the centers in these areas, we apply a learnable affine transformation to convert $S_m$ into a center-oriented map $\widetilde{S}$:

\begin{equation}
\widetilde{S} = W \odot S + B,
\end{equation}
where $W, B \in \mathbb{R}^{h\times w}$ denote learnable weights and biases.
We then concatenate $F_v$ with $\widetilde{S}$ to obtain a fused feature representation $F_0$. The fused features is subsequently processed by a sequence of point-wise convolution layers, progressively reducing the channel dimensionality. Finally, a lightweight decoding head consisting of $1\times1$Conv, normalization, GELU activation, $1\times1$Conv generates the instance center prediction map $\hat{M} \in \mathbb{R}^{h \times w}$, where each element indicates the probability of the corresponding spatial location being an instance center.
The feature projection process is formulated as:

\begin{equation}
F_{i+1} =
\phi_i\Big(
\mathrm{Norm}_i
\big(
\mathrm{Conv}(F_i)
\big)
\Big),
\end{equation}
where $\mathrm{Conv}(\cdot)$ denotes a point-wise convolution, $\mathrm{Norm}_i(\cdot)$ represents the normalization operation, and $\phi_i(\cdot)$ denotes the GELU activation function.

\subsection{Loss}

We follow the training losses of InstanceVG\cite{dai2025improving} for both instance detection and segmentation.  
For the detection branch, $\mathcal{L}_{\text{det}}$ adopts a DETR-style \cite{carion2020end} loss function, consisting of L1 loss, cross-entropy loss, and GIoU loss.  
For the segmentation branch, we employ a combination of binary cross-entropy (BCE) and Dice losses \cite{cheng2022masked}, and additionally apply similar supervision to the global semantic segmentation output. 
The Exist Head distinguishes the presence of targets by applying an MLP over average-pooled global semantic features $M_g$, and is optimized using a BCE loss.  
Moreover, we introduce auxiliary supervision for both the similarity map and the instance center prediction. $\mathcal{L}_{\text{sim}}$ supervises $S_m$ using the ground-truth mask, and $\mathcal{L}_{\text{icdm}}$ supervises $\hat{M}$ using the instance center density map $D$. The overall training objective is defined as:

\begin{equation}
\mathcal{L} = \mathcal{L}_{\text{det}} + \mathcal{L}_{\text{seg}} + \mathcal{L}_{\text{exist}} +\alpha\mathcal{L}_{\text{sim}} + \beta\mathcal{L}_{\text{icdm}},
\end{equation}
where $\alpha$ and $\beta$ are weighting coefficients that control the contribution of both auxiliary supervisions. For experiments on other hyperparameters in the loss function, see Appendix C.

\begin{table*}[!htbp]
\centering
\begin{minipage}[t]{0.58\textwidth}
\centering
\captionsetup{position=above}
\captionof{table}{Comparison with state-of-the-art methods on the R-RefCOCO/+/g \cite{wu2024toward} datasets.}
\resizebox{\linewidth}{!}{
\begin{tabular}{*{10}{c}}
     \toprule
     \multirow{2}*{Models} & \multicolumn{3}{c}{R-RefCOCO} & \multicolumn{3}{c}{R-RefCOCO+} & \multicolumn{3}{c}{R-RefCOCOg}   \\ 
        \cmidrule(lr){2-4} \cmidrule(lr){5-7} \cmidrule(lr){8-10} 
        & mIoU & mRR & rIoU & mIoU & mRR & rIoU & mIoU & mRR & rIoU \\
     \midrule
        CRIS \cite{liu2023caris} & 43.58 & 76.62 & 29.01 & 32.13 & 72.67 & 21.42 & 27.82 & 74.47 & 14.60 \\
        EFN \cite{feng2021encoder} & 58.33 & 64.64 & 32.53 & 37.74 & 77.12 & 24.24 & 32.53 & 75.33 & 19.44 \\
        VLT \cite{ding2021vision} & 61.66 & 63.36 & 34.05 & 50.15 & 75.37 & 34.19 & 49.67 & 67.31 & 31.64 \\
        LAVT \cite{yang2022lavt} & 69.59 & 58.25 & 36.20 & 56.99 & 73.45 & 36.98 & 59.52 & 61.60 & 34.91 \\
        LAVT+ \cite{yang2022lavt} & 54.70 & 82.39 & 40.11 & 45.99 & 86.35 & 39.71 & 47.22 & 81.45 & 35.46 \\
        RefSegformer \cite{wu2024toward} & 68.78 & 73.73 & 46.08 & 55.82 & 81.23 & 42.14 & 54.99 & 71.31 & 37.65 \\
        CoHD \cite{luo2025cohd} & 74.16 & 84.27 & 53.61 & 64.59 & 87.49 & 49.07 & 63.56 & 82.68 & 42.16 \\
        InstanceVG \cite{dai2025improving} & \uline{76.73} & \uline{92.15} & \uline{62.41} & \uline{69.73} & \textbf{94.63} & \uline{59.13} & \uline{70.16} & \uline{92.30} & \uline{54.36} \\
    \midrule
        Ours & \textbf{78.26} & \textbf{92.29} & \textbf{64.42} & \textbf{72.15} & \uline{94.20} & \textbf{61.14} & \textbf{72.40} & \textbf{92.67} & \textbf{57.42} \\
    \bottomrule
\end{tabular}
}
\label{tab_4}
\end{minipage}
\hfill
\begin{minipage}[t]{0.4\textwidth}
\centering
\captionsetup{position=above}
\captionof{table}{Comparison with state-of-the-art methods on the Ref-ZOM \cite{hu2023beyond} dataset.}
    \resizebox{\linewidth}{!}{
    \begin{tabular}{*{5}{c}}
         \toprule
         Models & Backbone & mIoU & oIoU & Acc.  \\
         \midrule
            \multicolumn{5}{c}{MLLM Methods} \\
            \midrule
             LISA-V-7B \cite{lai2024lisa} & SAM-ViT-H & 65.39 & 66.41 & 93.39 \\
            
             GSVA-V-7B \cite{xia2024gsva} & SAM-ViT-H & 68.13 & 68.29 & 94.59 \\
        \midrule
            \multicolumn{5}{c}{Specialist Methods} \\
            \midrule
            MCN \cite{luo2020multi} & DarkNet-53/GRU & 54.70 & 55.03 & 75.81 \\
            VLT \cite{ding2021vision} & DarkNet-56/RNN & 60.43 & 60.21 & 79.26 \\
            LAVT \cite{yang2022lavt} & Swin-B/BERT & 64.78 & 64.45 & 83.11 \\
            DMMI \cite{hu2023beyond} & Swin-B/BERT & 68.21 & 68.77 & 87.02 \\
            CoHD \cite{luo2025cohd} & Swin-B/BERT & 69.81 & 68.99 & 93.34 \\
            InstanceVG \cite{dai2025improving} & BEiT3-ViT-B & \uline{71.52} & \uline{71.12} & \uline{97.42} \\
            \midrule
            Ours & BEiT3-ViT-B & \textbf{73.28} & \textbf{72.78} & \textbf{97.98} \\
        \bottomrule
    \end{tabular}
    }
\label{tab_5}
\end{minipage}

\vspace{0.5em}

\begin{minipage}[t]{0.39\textwidth}
\centering
\captionsetup{position=above}
\captionof{table}{Ablation study on proposed modules.}
   \resizebox{\linewidth}{!}{
    \begin{tabular}{*{11}{c}}
     \toprule
     \multirow{2}*{Multi-task} & \multirow{2}*{SeDE} & \multirow{2}*{SpDE} & \multicolumn{4}{c}{Val} & \multicolumn{4}{c}{TestB} \\ 
        \cmidrule(lr){4-7} \cmidrule(lr){8-11} 
        & & & F1score & N-acc. & gIoU & cIoU & F1score & N-acc. & gIoU & cIoU \\
     \midrule
        & & & 66.28	& 58.34 & - & - & 57.53 & 57.52 & - & - \\
        
        \ding{51} & & & 69.32 & 63.71 & 70.45 & 69.32 & 58.63 & 59.81 & 65.55 & 65.31  \\

        \ding{51} & \ding{51} & & 70.35 & 65.02 & 70.8 & 69.46 & 59.41 & 61.24 & 65.73 & 65.42  \\

        \ding{51} & \ding{51} & \ding{51} & \textbf{73.45} & \textbf{74.85} &  \textbf{74.73} & \textbf{70.16} & \textbf{60.53} & \textbf{66.64} & \textbf{67.66} & \textbf{66.27} \\
     \bottomrule
\end{tabular}}
\label{tab_8}
\end{minipage}
\hfill
\begin{minipage}[t]{0.29\textwidth}
    \centering
    \captionsetup{position=above}
    \captionof{table}{Ablation study within SeDE.}
    \resizebox{\linewidth}{!}{
    \begin{tabular}{*{5}{c}}
     \toprule
        Ablations & F1score & N-acc. & gIoU & cIoU \\
     \midrule
        No $A_q$ & 71.17 & 72.60 & 73.99 & 69.89   \\
        
        CA & 72.65 & 73.05 & 74.09 & 69.90  \\

        SGCA/m & 73.55 & 73.94 & 74.32 & 70.07  \\

        SGCA & \textbf{73.45} & \textbf{74.85} & \textbf{74.73} & \textbf{70.16}  \\
     \bottomrule
\end{tabular}
    }
\label{tab_9}
\end{minipage}
\hfill
\begin{minipage}[t]{0.29\textwidth}
    \centering
    \captionsetup{position=above}
    \captionof{table}{Ablation study within SpDE.}
    \resizebox{\linewidth}{!}{
    \begin{tabular}{*{6}{c}}
     \toprule
     $S_m$ & $ \hat{M} $ & F1score & N-acc. & gIoU & cIoU \\
     \midrule
        & & 72.58 & 72.58 & 73.44 & 69.28  \\
        
        \ding{51} & & 72.83 & 73.1 & 73.97 & 69.91  \\

        & \ding{51} & 73.12 & 73.7 & 74.03 & 69.81  \\

        \ding{51} & \ding{51} & \textbf{73.45} & \textbf{74.85}  & \textbf{74.73} & \textbf{70.16}  \\
     \bottomrule
\end{tabular}
    }
\label{tab_10}
\end{minipage}

\end{table*}

\section{Experiments}

\subsection{Experimental Setup}

For REC and RES tasks, we conduct joint training on the RefCOCO \cite{yu2016modeling}, RefCOCO+ \cite{yu2016modeling}, and RefCOCOg \cite{nagaraja2016modeling,mao2016generation} datasets, and report evaluation results on each subset. For GREC and GRES tasks, training and evaluation are performed on the gRefCOCO \cite{liu2023gres,he2023grec}, Ref-ZoM \cite{hu2023beyond}, and R-RefCOCO/+/g \cite{wu2024toward} datasets, respectively. SSDE is built upon a pre-trained multi-modality encoder BEiT-3 with ViT-B, and the number of queries is set to 1 and 10 for the CVG and GVG tasks, and SSDE is trained for 25 and 10 epochs, respectively.

\subsection{Comparison with SOTA Methods}

\textbf{Classic visual grounding.}
Table \ref{tab_3} presents the performance of SSDE on the RES task. Our model surpasses both SAM-based \cite{kirillov2023segment} and LLM-based \cite{touvron2023llama} methods, such as GLaMM-7B \cite{rasheed2024glamm}, LISA-13B \cite{lai2024lisa}, and GSVA-13B \cite{xia2024gsva}, all of which rely on substantially larger model scales and are trained with more extensive grounding data. For specialist methods, compared with EEVG \cite{chen2024efficient} and OneRef \cite{xiao2024oneref}, which leverage additional datasets for pretraining (e.g., ReferIt \cite{kazemzadeh2014referitgame} or Flickr30k \cite{young2014image}), SSDE achieves superior performance across all evaluation benchmarks, with average improvements of up to 2.4\%, 5.7\%, and 3.9\% on the three split sets, respectively.
Under the same data scale, SSDE improves upon the state-of-the-art model RaAM-RVG \cite{ouyang2025region}, which adopts separate modality encoders, by an average of 2.2\%, 7.9\%, and 5.7\%, and also outperforms InstanceVG \cite{dai2025improving}, which
is the best model based on unified multi-modality encoder, by an average of 0.4\%, 1.4\%, and 1.2\% across the three splits, respectively.
Table \ref{tab_2} in Appendix reports the performance on the REC task. 
Overall, as a unified multi-task framework, SSDE consistently achieves superior performance over existing state-of-the-art methods on both REC and RES tasks.

\textbf{Generalized visual grounding.}
To further evaluate the effectiveness of SSDE under generalized settings, we first compare it with SOTA methods on the gRefCOCO \cite{liu2023gres} dataset, as shown in Table \ref{tab_6}. The results demonstrate that SSDE achieves state-of-the-art performance on GRES task. Compared with RaAM-RVG \cite{ouyang2025region}, SSDE improves average performance across the three split sets by 3.8\%, 2.5\%, and 1.7\%, respectively. When compared with LatentVG \cite{yu2025latent}, which adopts the same backbone with SSDE, the average improvements are 2.1\%, 1.9\%, and 1.7\%.

Furthermore, we extend our evaluation to the Ref-ZOM \cite{hu2023beyond} benchmark, as shown in Table \ref{tab_5}. SSDE consistently outperforms prior approaches, achieving gains of 1.76\% and 1.66\% in mIoU and oIoU over InstanceVG \cite{dai2025improving} (using the same backbone) and an average improvement of 3.97\% over COHD \cite{luo2025cohd}, a SOTA method based on independent modality encoding. We also report results on the R-RefCOCO \cite{wu2024toward} datasets in Table \ref{tab_4}, where SSDE maintains leading performance, surpassing COHD \cite{luo2025cohd} by an average margins of 7.6\%, 8.8\%, and 11.4\% across three different splits. Notably, our method also exceeds LLM-based methods trained on large-scale data.

Beyond segmentation, we further evaluate detection performance on the GREC task, as shown in Table \ref{tab_7}. Compared with HieA2G \cite{wang2025hierarchical}, a SOTA method based on independently encoded modalities, SSDE achieves average improvements of 11.0\%, 9.0\%, and 7.7\% across the three splits. Compared with PropVG \cite{dai2025propvg}, which uses the same backbone, SSDE still delivers consistent average gains of 2.6\%, 2.7\%, and 2.0\%. Overall, these results demonstrate that SSDE achieves consistently superior performance across GVG tasks.

\subsection{Ablation studies}

\textbf{Effects of each proposed component.}
Table \ref{tab_8} represents an analysis of the effectiveness of core modules. Both the SeDE and SpDE modules leads to notable performance improvements, while their combination yields the best results. This indicates that SpDE leverages instance center density map as explicit supervision, enabling to capture structured spatial distribution of multiple instances. Meanwhile, SeDE enhances query representations with  visual attributes, significantly improving instance-level discriminability. The joint modeling of structured spatial distribution and fine-grained semantic attributes ultimately results in substantial performance gains.

\begin{table}[t]
\centering
\caption{Comparison with state-of-the-art methods on the gRefCOCO \cite{liu2023gres} dataset for the GRES task.}
\resizebox{\linewidth}{!}{
    \begin{tabular}{*{8}{c}}
     \toprule
     \multirow{2}*{Models} & \multirow{2}*{Backbone} & \multicolumn{2}{c}{Val} & \multicolumn{2}{c}{TestA} & \multicolumn{2}{c}{TestB} \\ 
        \cmidrule(lr){3-4} \cmidrule(lr){5-6} \cmidrule(lr){7-8} 
        & & gIoU & cIoU & gIoU & cIoU & gIoU & cIoU \\
     \midrule
        \multicolumn{8}{c}{MLLM Methods} \\
        \midrule
        LISA-V-7B \cite{lai2024lisa} & SAM-ViT-H & 61.63 & 61.76 & 66.27 & 68.50 & 58.84 & 60.63 \\
        
        GSVA-V-7B \cite{xia2024gsva} & SAM-ViT-H & 66.47 & 63.29 & 71.08 & 69.93 & 62.23 & 60.47 \\
 
    \midrule
        \multicolumn{8}{c}{Specialist Methods} \\
        \midrule
        MattNet \cite{yu2018mattnet} & ResNet-101/LSTM & 48.24 & 47.51 & 59.30 & 58.66 & 46.14 & 45.33 \\
        
        CRIS \cite{liu2023caris} & CLIP-R101/CLIP & 56.27 & 55.34 & 63.42 & 63.82 & 51.79 & 51.04 \\

        ReLA \cite{liu2023gres} & Swin-B/BERT & 63.60 & 62.42 & 70.03 & 69.26 & 61.02 & 59.88 \\
        
        HieA2G \cite{wang2025hierarchical} & Swin-B/RoBETRa & 68.40 & 64.20 & 72.00 & 70.40 & 62.80 & 61.00 \\

        COHD \cite{luo2025cohd} & Swin-B/BERT & 68.42 & 65.17 & 72.67 & 71.85 & 63.60 & 62.63 \\

        RaAM-RVG \cite{ouyang2025region}& ViT-B/BERT & 70.02 & 67.35 & 73.86 & 72.98 & 65.77 & 64.34 \\
    \midrule
        Latent-VG \cite{yu2025latent} & BEiT3-ViT-B & 72.45 & 68.23 & 74.51 & 73.53 & 66.12 & 64.16 \\

        PropVG \cite{dai2025propvg} & BEiT3-ViT-B & 73.29 & \uline{69.23} & 74.43 & 74.20 & 65.87 & 64.76 \\

        InstanceVG \cite{dai2025improving}& BEiT3-ViT-B & \uline{73.36} & 69.22 & \uline{75.21} & \uline{74.51} & \uline{66.74} & \uline{65.67} \\
    \midrule
        Ours & BEiT3-ViT-B & \textbf{74.83} & \textbf{70.09} & \textbf{76.23} & \textbf{75.51} & \textbf{67.45} & \textbf{66.13} \\
     \bottomrule
\end{tabular}
}
\label{tab_6}
\end{table}

\begin{table}[!htbp]
\centering
\caption{Comparison with state-of-the-art methods on the gRefCOCO \cite{he2023grec} dataset for the GREC task.}
\resizebox{\linewidth}{!}{
    \begin{tabular}{*{8}{c}}
     \toprule
     \multirow{2}*{Models} & \multirow{2}*{Backbone} & \multicolumn{2}{c}{Val} & \multicolumn{2}{c}{TestA} & \multicolumn{2}{c}{TestB} \\ 
        \cmidrule(lr){3-4} \cmidrule(lr){5-6} \cmidrule(lr){7-8} 
        & & F1score & N-acc. & F1score & N-acc. & F1score & N-acc. \\
     \midrule
        VLT \cite{ding2021vision}& DarkNet-56/RNN & 36.6 & 35.2 & 40.2 & 34.1 & 30.2 & 32.5 \\
        
        MDETR \cite{kamath2021mdetr} & ResNet-101/RoBERTa & 42.7 & 36.3 & 50.0 & 34.5 & 36.5 & 31.0 \\

        UNINEXT \cite{yan2023universal} & ResNet-50/BERT & 58.2 & 50.6 & 46.4 & 49.3 & 42.9 & 48.2 \\

        SimVG \cite{dai2024simvg} & BEiT3-ViT-B & 62.1 & 54.7 & 64.6 & 57.2 & 54.8 & 57.2  \\
        
        HieA2G \cite{wang2025hierarchical} & R101/RoBETRa & 67.8 & 60.3 & 66.0 & 60.1 & 56.5 & 56.0 \\

        PropVG \cite{dai2025propvg} & BEiT3-ViT-B & \uline{72.2} & \uline{72.8} & \uline{68.8} & \uline{69.9} & \uline{59.0} & \uline{65.0} \\

    \midrule
        Ours & BEiT3-ViT-B & \textbf{74.17} & \textbf{75.93} & \textbf{71.19} & \textbf{72.93} & \textbf{60.62} & \textbf{67.28}  \\
     \bottomrule
\end{tabular}
}
\label{tab_7}
\end{table} 

\textbf{Ablation study within SeDE.}
Table \ref{tab_9} presents the ablation study within the SeDE module. Firstly, compared to using only the subject-level semantics derived from textual expression as queries, incorporating visual attributes leads to significant improvements, particularly in terms of F1score and N-acc. This demonstrates the importance of fine-grained attribute modeling in enhancing the discriminability of target instances. Moreover, the strategy for extracting visual attributes has a critical role in overall performance. We evaluate three variants: (1) standard cross-attention, (2) SGCA without the scaling factor $m$, and (3) SGCA with the scaling factor $m$. The results indicate that the third variant achieves the best performance. When only standard cross-attention is employed, the learned attribute representation $A_q$ lacks explicit semantic constraints, making it susceptible to background noise. Introducing the similarity map as an attention bias (SGCA without $m$) injects spatial priors to some extent, but its effectiveness remains limited due to the lack of proper scale regulation. By further incorporating the scaling factor $m$, the influence of spatial priors can be regulated, leading to more reliable and target-relevant attribute extraction, thereby yielding superior performance.

\textbf{Ablation study within SpDE.}
Table \ref{tab_10} reports the ablation results within the SpDE module. 
Compared to learning similarity maps solely, modeling instance-centered density maps yields inferior results. This suggests that the density map provides a more stable and structured representation of the spatial distributions, where each instance corresponds to a distinct local peak. In contrast, the similarity map tends to cover the entire target region, lacking the ability to distinguish individual instances and thus failing to provide clear structural guidance for the spatial distribution in multi-instance scenarios. Furthermore, during the learning of the instance center density map, the supervised $S_m$ introduces more reliable spatial priors, further improving overall performance.

\begin{figure}[t]
  \centering
  \includegraphics[width=\linewidth]{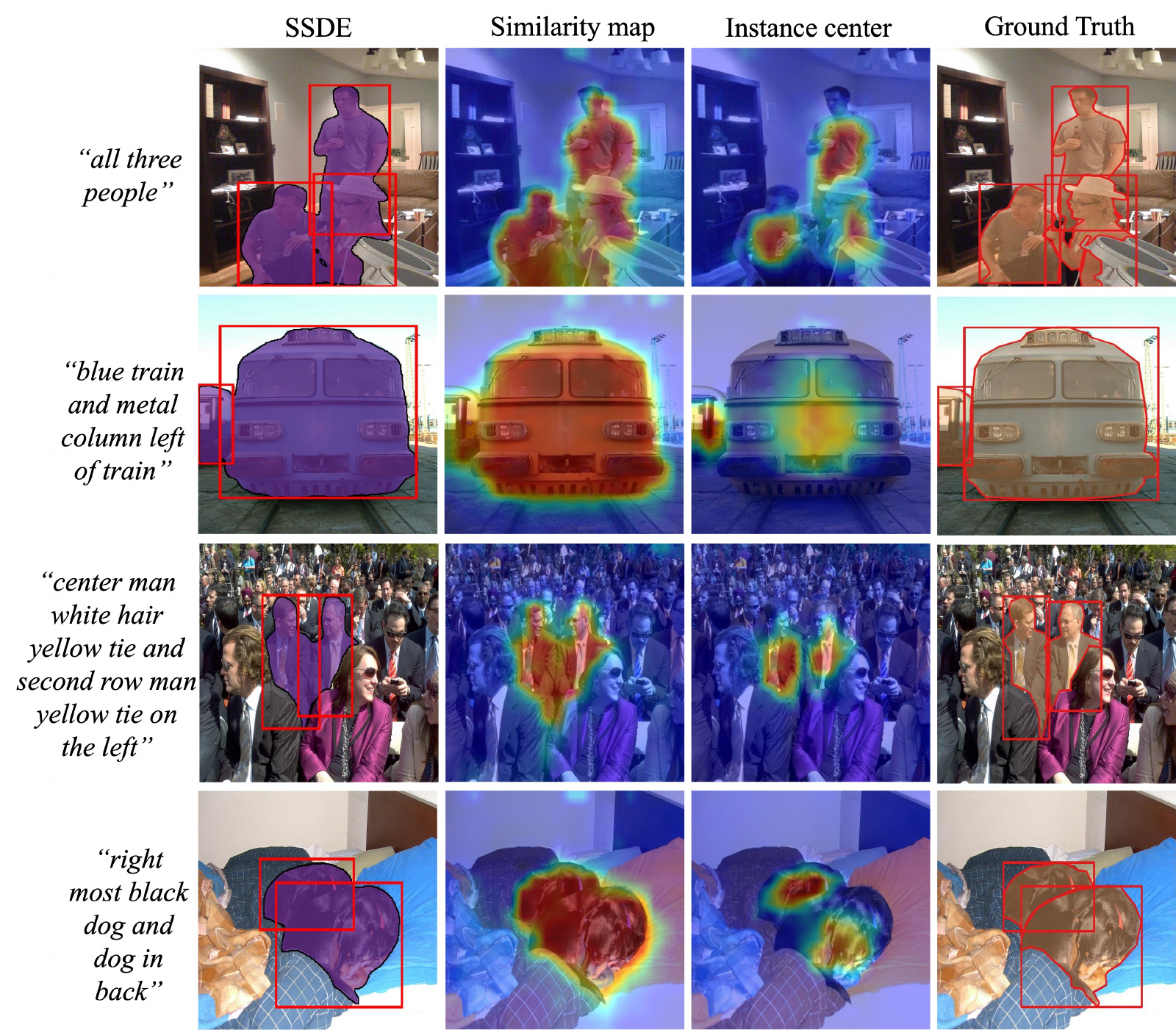}
  \caption{Visualizations of the prediction results, similarity maps, and instance centers predicted by the proposed method.}
  \label{fig_3}
\end{figure}

\subsection{Visualization}

In the Figure \ref{fig_3}, we visualize the prediction results, cross-modal similarity maps, instance center density maps, and ground truth for several multi-instance scenarios from the gRefCOCO \cite{liu2023gres} dataset. The visualizations demonstrate that SSDE achieves accurate cross-modal understanding, with the similarity maps effectively covering all targets. In challenging cases involving overlapping or adjacent instances, the instance center maps exhibits clear spatial separability, effectively distinguishing individual instances.
By explicitly predicting instance centers, SSDE implicitly encodes clearly spatial distribution, enabling counting information to be naturally integrated into the localization process and thereby improving multi-instance reasoning.

\section{Conclusion}

In this paper, we present SSDE for generalized visual grounding, aimed at improving cross-modal understanding and instance-level localization by jointly enhancing semantic and spatial discriminability. Specifically, we propose a semantic discriminability enhancement module that incorporates fine-grained visual attributes via spatially guided cross-modal interaction, effectively enlarging the inter-instance discriminative margins and mitigating omissions or ambiguities caused by insufficient visual cues. We also propose a spatial discriminability enhancement module that explicitly models instance center distributions, imposing instance-level structural constraints through auxiliary supervision, while also providing region-level guidance via instance-level supervision, thereby promoting the separability and independence of multiple targets.
Experimental results demonstrate that SSDE achieves state-of-the-art performance on multiple visual grounding benchmarks.

\begin{acks}
This work was supported by CAS Project for Young Scientists in Basic Research (YSBR-083) and Lingang Laboratory (Grant No. LGL-2616-04).
\end{acks}

\bibliographystyle{ACM-Reference-Format}
\bibliography{samples/sample-base}


\clearpage

\appendix

\begin{strip}
\begin{center}
    {\fontsize{18}{18}\selectfont \textbf{Semantic-Spatial Discriminability Enhancement for Generalized Visual Grounding} \\[0.5em] {\fontsize{14}{12}\selectfont Supplementary Material}}\\[0.5em]
\end{center}
\end{strip}

\section{Datasets}

\textbf{RefCOCO and RefCOCO+} \cite{yu2016modeling} are constructed using a two-player interactive annotation protocol similar to ReferItGame. RefCOCO contains 142,209 referring expressions for 50,000 objects across 19,994 images, while RefCOCO+ has a comparable scale, comprising 141,564 expressions for 49,856 objects in 19,992 images. Both datasets are split into training, validation, and test sets, with the test set further divided into testA and testB. Specifically, testA contains images with multiple people, whereas testB focuses on images with multiple instances of other object categories. Unlike RefCOCO, RefCOCO+ explicitly prohibits the use of location-related terms (e.g., “left”, “right”) and instead provides expressions based on object attributes, making it more challenging.

\textbf{RefCOCOg} \cite{mao2016generation,nagaraja2016modeling} is collected in a non-interactive manner via Amazon Mechanical Turk (AMT), resulting in longer and more complex expressions. Compared to RefCOCO and RefCOCO+, whose expressions are relatively concise (with average lengths of 3.61 and 3.53 words, respectively), RefCOCOg exhibits an average length of 8.4 words, reflecting richer semantic content and contextual dependencies. The dataset contains 85,474 referring expressions for 54,822 objects across 26,711 images. Following prior work, we adopt the UMD split, which divides the data into training, validation, and test sets without overlap between training and validation images.

\textbf{gRefCOCO} \cite{liu2023gres} extends RefCOCO by incorporating multi-target and non-target expressions for the GRES task. It contains 278,232 referring expressions, including 80,022 multi-target and 32,202 non-target expressions, with the remainder corresponding to single-target cases. The dataset covers 60,287 object instances across 19,994 images and follows the UNC partition, consisting of training, validation, testA, and testB splits.

\textbf{Ref-ZOM} \cite{hu2023beyond} is derived from the COCO dataset and consists of 55,078 images with 74,942 annotated objects. Among them, 43,749 images and 58,356 objects are used for training, while 11,329 images and 16,586 objects are used for testing. The annotations cover three scenarios: one-to-zero, one-to-one, and one-to-many, corresponding to no-target, single-target, and multi-target settings in GRES, respectively.

\textbf{R-RefCOCO} \cite{wu2024toward} includes three subsets: R-RefCOCO, R-RefCOCO+, and R-RefCOCOg, all derived from the RES benchmarks RefCOCO/+/g. The validation set follows the UNC partition and is officially adopted for evaluation. It is constructed by augmenting the training set with negative expressions at a 1:1 ratio relative to positive ones.

\section{Implement Details}
\subsection{Metrics}

\textbf{Classical visual grounding (REC/RES).} For REC, we adopt Precision@0.5 as the evaluation metric, which measures the proportion of predictions whose Intersection-over-Union (IoU) with the ground truth exceeds 0.5. For RES, we use mIoU, computed as the mean IoU over all predictions.  

\textbf{Generalized visual grounding (GREC/GRES).} For GREC, we employ Pr@(F1=1, IoU$\geq$0.5) and N-acc. as the main metrics. Specifically, Pr@(F1=1, IoU$\geq$0.5), denoted as F1score, evaluates the performance using an F1score of 1 with an IoU threshold of 0.5. A predicted bounding box is considered a true positive (TP) if it overlaps with a ground-truth box with an IoU of at least 0.5. If multiple predictions match, only the one with the highest IoU is counted as TP.
For GRES, different metrics are adopted across datasets. We report gIoU and cIoU for gRefCOCO \cite{liu2023gres}. Specifically, gIoU computes the average IoU over all instances within an image, where no-target cases are treated as true positives with an IoU of 1. The cIoU metric measures the ratio between intersection pixels and union pixels.
For Ref-ZOM \cite{hu2023beyond}, we adopt Acc., oIoU, and mIoU as evaluation metrics, where mIoU calculates the average IoU for images containing referred objects and oIoU corresponds to cIoU. For R-RefCOCO \cite{wu2024toward}, we report mIoU, mRR, and rIoU. The rIoU metric evaluates segmentation quality by incorporating negative sentences and assigning equal weight to positive instances in the mIoU calculation, while mRR measures the recognition rate of non-target expressions.

\subsection{Experiment setup}

Our model is built upon a pre-trained BEiT-3 multi-modality encoder with a ViT-B model size, featuring a patch size of 16 and a channel dimension of 768. We initialize the multi-modality encoder with the \texttt{beit3\_base\_indomain\_patch16\_224} pre-trained weights. Visual features from the 4th, 8th, and 12th layers of the backbone are extracted and fed into a feature pyramid network (FPN) \cite{li2022exploring} to obtain multi-scale representations.  
For CVG and GVG tasks, the maximum text lengths are set to 20 and 50, respectively, and the number of queries $N_q$ is set to 1 and 10. The models are trained for 25 fot CVG and 10 epochs for GVG. The loss weights $ \alpha$ and $ \beta$ for $\mathcal{L}_{\text{sim}}$ and $\mathcal{L}_{\text{icdm}}$ are both set to 0.1.  

We follow the training protocol of InstanceVG \cite{dai2025improving}, where the learning rate of the multi-modality encoder is set to $5\times10^{-5}$, and the remaining parameters are optimized with a learning rate of $5\times10^{-4}$. The model is trained using the Adam optimizer with full-precision training. The input image resolution is $320\times320$, and the batch size is set to 48. All experiments are conducted on a single NVIDIA A6000 GPU.

\begin{table*}[!htbp]
\caption{Comparison with state-of-the-art methods on the RefCOCO/+/g \cite{yu2016modeling} datasets for the REC task. In the “Backbone” column, a “/” indicates separate backbones for visual and textual modalities. “FT” indicates whether fine-tuning is performed on the target dataset, and “MT” denotes whether the model is trained in a multi-task setting. Precision@0.5 is used as the evaluation metric. The best results are highlighted in bold, and the second-best results are underlined.}
\centering
\resizebox{\linewidth}{!}{
    \begin{tabular}{*{13}{c}}
        \toprule
        \multirow{2}*{Models} & \multirow{2}*{Publication} & \multirow{2}*{Backbone} & \multirow{2}*{FT} & \multirow{2}*{MT} & \multicolumn{3}{c}{RefCOCO} & \multicolumn{3}{c}{RefCOCO+} & \multicolumn{2}{c}{RefCOCOg} \\ 
        \cmidrule(lr){6-8} \cmidrule(lr){9-11} \cmidrule(lr){12-13} 
        & & & & & val & testA & testB & val & testA & testB & val(u) & test(u) \\
      
        \midrule
        \multicolumn{13}{c}{MLLM Methods} \\
        \midrule
        Shikra-7B \cite{chen2023shikra} & arxiv23 & CLIP-ViT-L & \ding{51} & \ding{55} & 87.01 & 90.61 & 80.24 & 81.60 & 87.36 & 72.12 & 82.27 & 82.19 \\
        
        LISA-L2-13B \cite{lai2024lisa} & CVPR24 & SAM-ViT-H & \ding{51} & \ding{55} & 85.91 & 88.84 & 81.73 & 74.46 & 80.56 & 68.26 & 80.09 & 81.17 \\
        
        GSVA-L2-13B \cite{xia2024gsva} & CVPR24 & SAM-ViT-H & \ding{51} & \ding{55} & 89.16 & 92.08 & 87.17 & 79.74 & 84.45 & 73.41 & 85.47 & 86.18 \\
        \midrule
        \multicolumn{13}{c}{Specialist Methods} \\
        \midrule
        SeqTR \cite{zhu2022seqtr} & ECCV22 & DarkNet-53/GRU & \ding{51} & \ding{51} & 87.00 & 90.15 & 83.59 & 78.69 & 84.51 & 71.87 & 82.69 & 83.37 \\
        
        GroundingDINO \cite{liu2024grounding} & ECCV24 & Swin-T/BERT-B & \ding{51} & \ding{55} & 89.19 & 91.86 & 85.99 & 81.09 & 87.40 & 74.71 & 84.15 & 84.94 \\
        
        HiVG-B \cite{xiao2024hivg} & ACMMM24 & CLIP-B/CLIP-B & \ding{51} & \ding{55} & 90.56 & 92.55 & 87.23 & 83.08 & 87.83 & 76.68 & 84.71 & 84.69 \\
        
        ReMeREC \cite{hu2025remerec} & ACMMM25 & RN50-DETR/BERT-B & \ding{51} & \ding{55} & 89.63 & 91.91 & 86.56 & 84.31 & 86.29 & 78.89 & 86.76 & 87.30 \\
       
        HieA2G \cite{wang2025hierarchical} & AAAI25 & ResNet101/RoBERTa-B & \ding{51} & \ding{51} & 87.80 & 90.30 & 84.00 & 80.70 & 85.60 & 72.90 & 83.70 & 83.80 \\

        RaAM-RVG \cite{ouyang2025region} & ICCV25 & ViT-B/BERT-B & \ding{55} & \ding{51} & 91.45 & 93.42 & 88.71 & 84.78 & 89.23 & 79.24 & 86.78 & 87.71 \\
        \midrule
        SimVG-DB \cite{dai2024simvg} & NerulPS24 & BEiT3-ViT-B & \ding{51} & \ding{55} & 91.47 & 93.65 & 87.94 & 84.83 & 88.85 & 79.12 & 86.30 & 87.26 \\

        OneRef-B \cite{xiao2024oneref} & NerulPS25 & BEiT3-ViT-B & \ding{51} & \ding{51} & \uline{91.89} & 94.31 & 88.58 & 86.38 & 90.38 & 79.47 & 86.82 & \uline{87.32} \\
 
        Latent-VG \cite{yu2025latent} & ICCV25 & BEiT3-ViT-B & \ding{55} & \ding{51} & 91.75 & \uline{94.64} & \uline{88.62} & \uline{86.41} & \uline{90.57} & \uline{80.59} & \uline{87.01} & 87.11 \\

        PropVG \cite{dai2025propvg} & ICCV25 & BEiT3-ViT-B & \ding{55} & \ding{51} & 88.96 & 91.55 & 85.73 & 83.72 & 88.00 & 76.60 & 83.50 & 84.44\\
        \midrule
        Ours & - & BEiT3-ViT-B & \ding{55} & \ding{51} & \textbf{92.55} & \textbf{94.66} & \textbf{89.44} & \textbf{87.43} & \textbf{91.48} & \textbf{82.76} & \textbf{88.62} & \textbf{88.70} \\
        \bottomrule
        
    \end{tabular}}
    \label{tab_2}
\end{table*}

\section{Additional Experiments}

\subsection{Results on REC}

Table \ref{tab_2} presents the performance on the REC task. Our model significantly outperforms SAM-based and LLM-based methods, such as Shikra-7B \cite{chen2023shikra}, LISA-13B \cite{lai2024lisa}, and GSVA-13B \cite{xia2024gsva}, all of which utilize more extensive grounding data and have larger model sizes.
Compared with task-specific models, SSDE consistently outperforms HieA2G \cite{wang2025hierarchical} and OneRef \cite{xiao2024oneref}, which are trained under a compositional setting using additional datasets. Specifically, our approach improves the average performance across the three split sets by up to 4.9\%, 7.5\%, and 4.9\%, respectively. Compared with methods trained at a similar data scale, SSDE consistently outperforms recent state-of-the-art methods, including those that adopt separate modality encoding, such as RaAM-RVG \cite{ouyang2025region}, as well as unified multi-modality encoder-based approaches, such as Latent-VG \cite{yu2025latent}.On the three different splits, our approach achieves average gains of up to 1.0\%, 2.8\%, and 1.6\%, respectively, demonstrating its strong effectiveness and robustness.

\subsection{Ablation Studies}

\textbf{Instance center determination.} 
In constructing the instance center density map, the strategy for determining instance centers plays a critical role in overall performance. Due to occlusions, a single object may produce multiple disconnected mask regions. Table \ref{tab_11} compares three strategies: (1) using the centers of all connected components, (2) using the center of the central component, and (3) the center of the largest connected component for each instance. The results indicate that the third strategy yields the best performance. The first strategy may assign multiple centers to a single instance, making it difficult for the model to learn a stable and consistent instance distribution. The second strategy is sensitive to the spatial distribution of fragmented regions and may introduce localization bias toward small local structures. In contrast, selecting the largest connected component provides a more reliable representation of the primary object region, thereby improving the accuracy of subsequent reference point initialization.

\begin{table}[!htbp]
\caption{Ablation study on methods for determining the center of an instance.}
\centering
    \begin{tabular}{*{5}{c}}
     \toprule
        Ablations & F1score & N-acc. & gIoU & cIoU \\
     \midrule
        All & 72.02 & 70.89 & 72.86 & 69.26    \\
        
        Nearest & 72.34 & 71.60 & 73.12 & 69.15  \\

        Largest & \textbf{73.45} & \textbf{74.85} & \textbf{74.73} & \textbf{70.16}  \\
     \bottomrule
\end{tabular}
\label{tab_11}
\end{table}

\begin{table}[!htbp]
\caption{Ablation study on $\alpha$ and $\beta$.}
\centering
\resizebox{\linewidth}{!}{
    \begin{tabular}{*{8}{c}}
     \toprule
        \multirow{2}*{$\alpha$} & \multirow{2}*{$\beta$} & \multicolumn{2}{c}{Val} & \multicolumn{2}{c}{TestA} & \multicolumn{2}{c}{TestB} \\ 
        \cmidrule(lr){3-4} \cmidrule(lr){5-6} \cmidrule(lr){7-8} 
        & & F1score & gIoU & F1score & gIoU & F1score & gIoU \\
     \midrule
        0.05 & 0.05 & \uline{73.62} & \uline{74.30} & \textbf{70.76}	& 76.01	& 59.94 & 66.99   \\
        
        0.05 & 0.1 & 71.99 & 73.94 & 69.55 & 75.74	&	59.63 & 67.4 \\

        0.1	& 0.05 & 73.31  & 74.28 & 70.29	& 75.62	& 	\uline{60.29} & \uline{67.23}  \\

        0.1 & 0.1 & 73.45 & \textbf{74.73} & \uline{70.64} & \uline{76.04} &	\textbf{60.53}	& \textbf{67.66}	 \\
        
        0.1 & 0.2 & \textbf{73.90}  & \textbf{74.73} & \textbf{70.76} & \textbf{76.08} & 59.97 & 66.83 \\
     \bottomrule
\end{tabular}}
\label{tab_12}
\end{table}

\begin{figure*}
\centering
\includegraphics[width=6.8in]{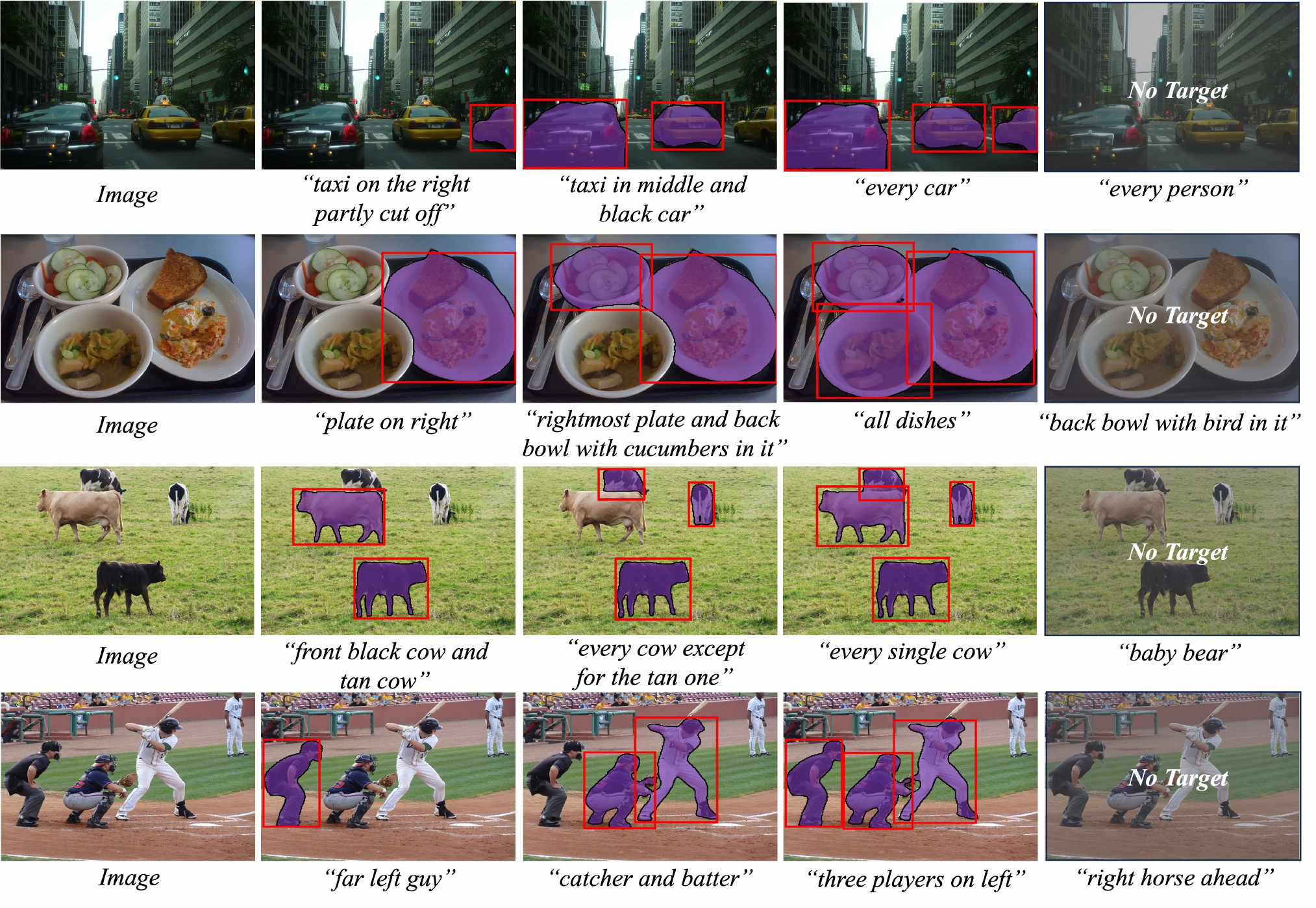}
\caption{Visualization on gRefCOCO \cite{liu2023gres} dataset.}
\label{fig_5}
\end{figure*}

\begin{table}[!htbp]
\caption{Ablation study on loss terms.}
\centering
\resizebox{\linewidth}{!}{
    \begin{tabular}{*{7}{c}}
     \toprule
     $\lambda_{det}$ & $\lambda_{seg}$ & $\lambda_{exist} $ & F1score & N-acc. & gIoU & cIoU \\ 
     \midrule
        0.1 & 1.0 & 0.2 & 73.45 & 74.85	& 74.73 & \textbf{70.16} \\
        
        0.2 & 1.0 & 0.2 & \textbf{74.17} & \textbf{75.93} & \textbf{74.83} & 70.09  \\
        0.1 & 1.0 & 0.3 & 72.96 & 72.6 & 73.81 & 69.7 \\
        0.2 & 1 & 0.1 & 74.02 & 74.65 & 74.29 & 69.62  \\
        0.2 & 1.5 & 0.2 & 73.69 & 75.08 & 74.81 & 69.99  \\

        0.2 & 0.5 & 0.2 & 73.24 & 73.13 & 73.96 & 69.63 \\
     \bottomrule
\end{tabular}}
\label{tab_15}
\end{table} 

\textbf{Hyperparameter in loss.}
We further analyze the impact of different values of $\alpha$ and $\beta$, as reported in Table \ref{tab_12}. The loss $\mathcal{L}_{\text{sim}}$ enforces cross-modal semantic alignment, while $\mathcal{L}_{\text{icdm}}$ models the spatial distribution of instance centers. The results indicate that the optimal performance is achieved when both $\alpha$ and $\beta$ are set to 0.1. This suggests that maintaining a balanced contribution of these auxiliary constraints is beneficial, enabling the model to jointly learn semantic relevance and instance-level structural information, thus yielding more stable performance gains in multi-instance scenarios.

SSDE includes DETR-style loss, instance segmentation loss, exist loss, and auxiliary structural regularization losses. The ablation of auxiliary loss is in Table \ref{tab_12}. The ablation of others is shown in Table \ref{tab_15}. After systematically tuning weights on gRefcoco, the final loss weights are set to 0.2, 1.0, 0.2, 0.1, and 0.1. This configuration achieves the optimal trade-off of grounding accuracy.

\begin{table}[!htbp]
\caption{Ablation study on the number of queries for GRES task.}
\centering
    \begin{tabular}{*{5}{c}}
     \toprule
       Ablations & F1score & N-acc. & gIoU & cIoU \\
     \midrule
        5 & \textbf{73.74} & 73.67 & 74.13 & 69.85 \\
        10 & 73.45 & \textbf{74.85}  & \textbf{74.73} & \textbf{70.16} \\
        
        20 & 73.39 & 73.8  & 74.29 & 70.03 \\

        30 & 72.07 & 72.01 & 73.76 & 69.71 \\
     \bottomrule
\end{tabular}
\label{tab_13}
\end{table}

\begin{figure*}
\centering
\includegraphics[width=6.9in]{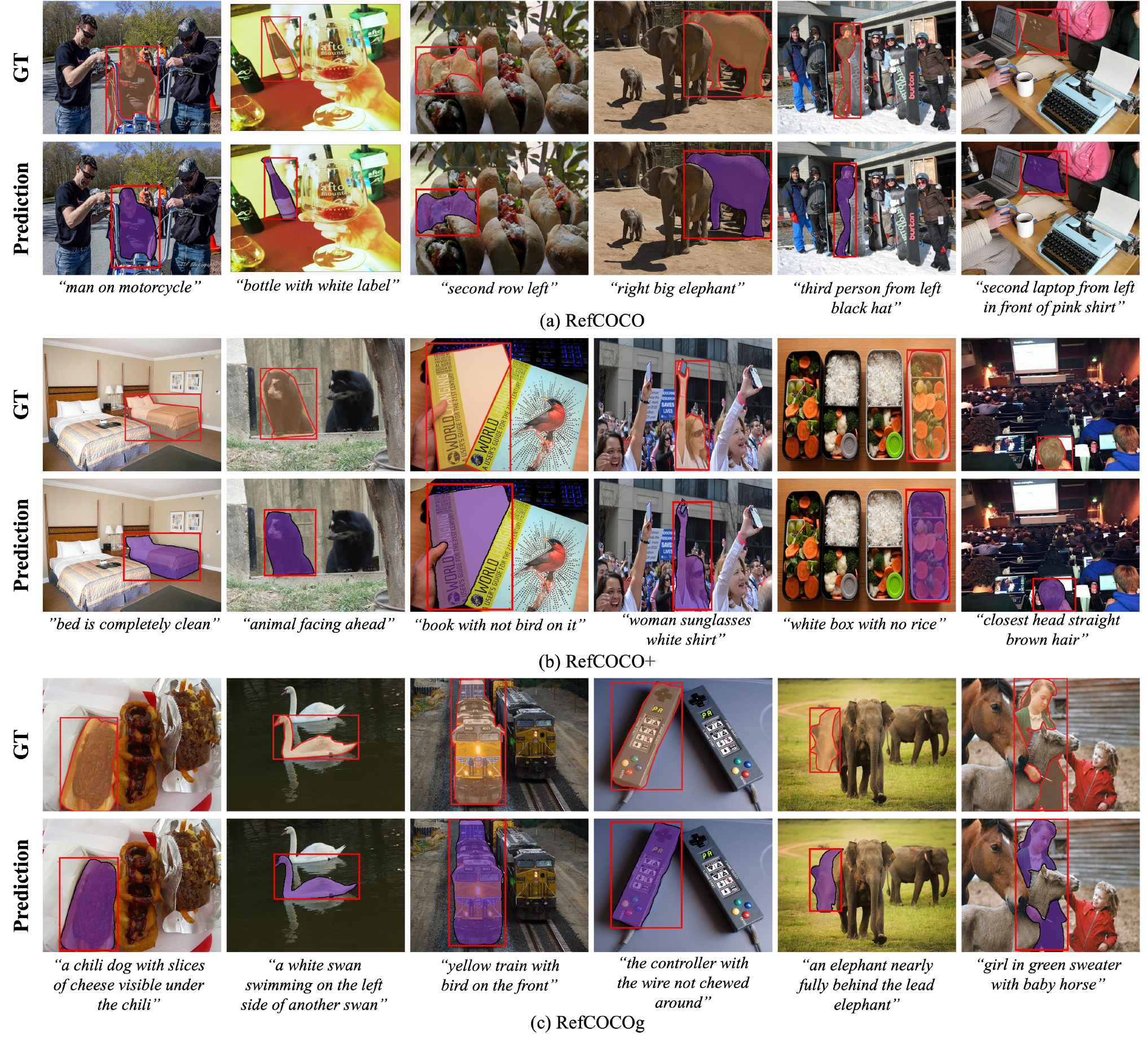}
\caption{Visualization on RefCOCO/+/g \cite{yu2016modeling} dataset.}
\label{fig_4}
\end{figure*}

\textbf{Query numbers for GRES.}
Table \ref{tab_13} reports the impact of varying the number of queries in the GRES task. 
The results show that setting the number of queries to 10 leads to the best performance, as it facilitates a more stable cross-modal coupling within the feature space. In contrast, using an excessive number of queries tends to dilute semantic focus, increase optimization difficulty, and introduce additional uncertainty, ultimately resulting in degraded performance.

\begin{table}[!htbp]
\caption{Efficiency and performance comparison with other methods on the testB set of RefCOCO+. “FT” indicates whether pre-training is performed, “MT” indicates whether it is multi-task training, and “Params” and “GFlops” represent the number of parameters and computational cost required by the models.}
\centering
\resizebox{\linewidth}{!}{
    \begin{tabular}{*{8}{c}}
     \toprule
       Methods & FT & MT & Backbone & Params(M) & GFlops & Prec & mIoU \\
     \midrule
        TransVG++ \cite{deng2023transvg++} & \ding{51} &\ding{55} & ViT-B/BERT & 206.66 & 396.00 & $-$ & 75.39 \\
        
        ReLA \cite{liu2023gres} & \ding{55} & \ding{55} & Swin-B/BERT & 226.00 & 131.00 & 57.65 & $-$ \\

        COHD \cite{luo2025cohd} & \ding{55} & \ding{55} & Swin-B/BERT	& 248.00 & 133.00 & 65.45 & $-$ \\

        One-Ref	\cite{xiao2024oneref} & \ding{51} & \ding{51} & BEiT3-ViT-B & 267.00  & 162.00 & 69.58 &	79.47  \\
        
        PropVG \cite{dai2025propvg} & \ding{55} & \ding{51} & BEiT3-ViT-B & 191.39 & 217.86 & 67.22	& 76.60  \\
        Latent-VG \cite{yu2025latent} & \ding{55} & \ding{51} & BEiT3-ViT-B & 267.00 &	198.00 & 72.95 &	80.59 \\
        InstanceVG \cite{dai2025improving} & \ding{55}	& \ding{51} & BEiT3-ViT-B & 182.69  & 118.84  & 71.56 & 80.70 \\
        SSDE & \ding{55} & \ding{51} & BEiT3-ViT-B & 183.85  & 119.34 & \textbf{74.00} & \textbf{82.76} \\
     \bottomrule
\end{tabular}}
\label{tab_14}
\end{table}

\subsection{Efficienty and parameters}

In the Table \ref{tab_14}, we compare the parameters and computational efficiency of SSDE with those of other visual grounding methods. We employ the \textit{thop} library to calculate the parameters and GFlops for SSDE, PropVG \cite{dai2025propvg}, and InstanceVG \cite{dai2025improving}, while the statistics for the remaining models are obtained from their respective publications. Overall, methods \cite{deng2023transvg++,liu2023gres,luo2025cohd} based on unimodal encoders typically exhibit larger model sizes compared to those adopting unified multimodal encoders \cite{dai2025improving,dai2025propvg}. SSDE achieves the best overall performance while maintaining a relatively low number of parameters and computational cost. Furthermore, 
compared to InstanceVG \cite{dai2025improving}, our method introduces only marginal increases in parameters and computational overhead, yet delivers substantially superior performance, demonstrating a more favorable efficiency–performance trade-off.

\begin{figure*}
\centering
\includegraphics[width=6.9in]{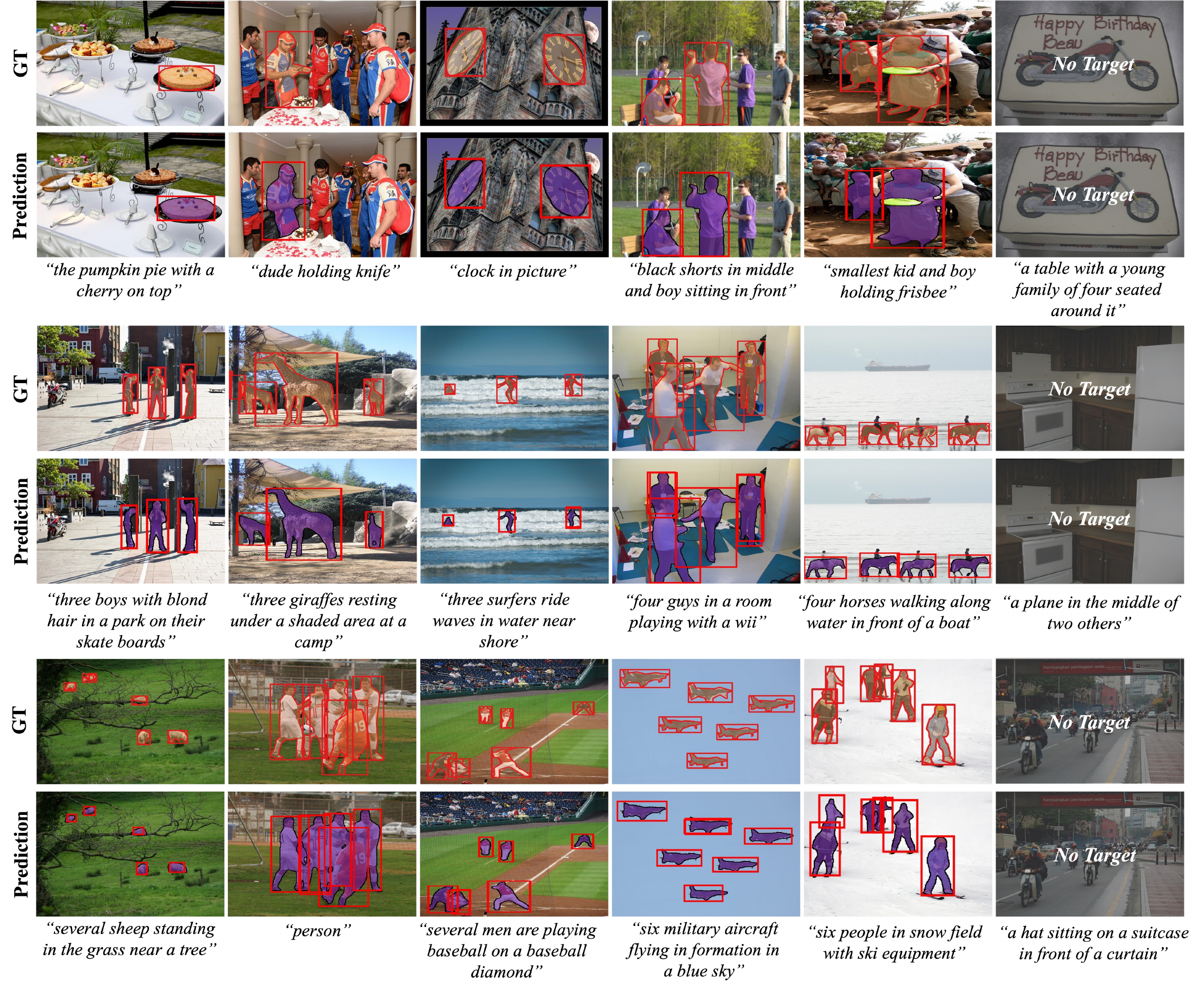}
\caption{Visualization on Ref-ZOM \cite{hu2023beyond} dataset.}
\label{fig_6}
\end{figure*}

\section{Visualization}

\textbf{Visualization on CVG task.}
We visualize representative predictions on RefCOCO/+/g \cite{yu2016modeling,nagaraja2016modeling,mao2016generation}, as illustrate in Figure \ref{fig_4}. The results demonstrate that SSDE accurately localizes the referred targets, even for long and complex expressions, demonstrating its strong fine-grained semantic understanding and the effectiveness of leveraging visual attributes to complement subject-level textual semantics.

\begin{figure*}
\centering
\includegraphics[width=6.9in]{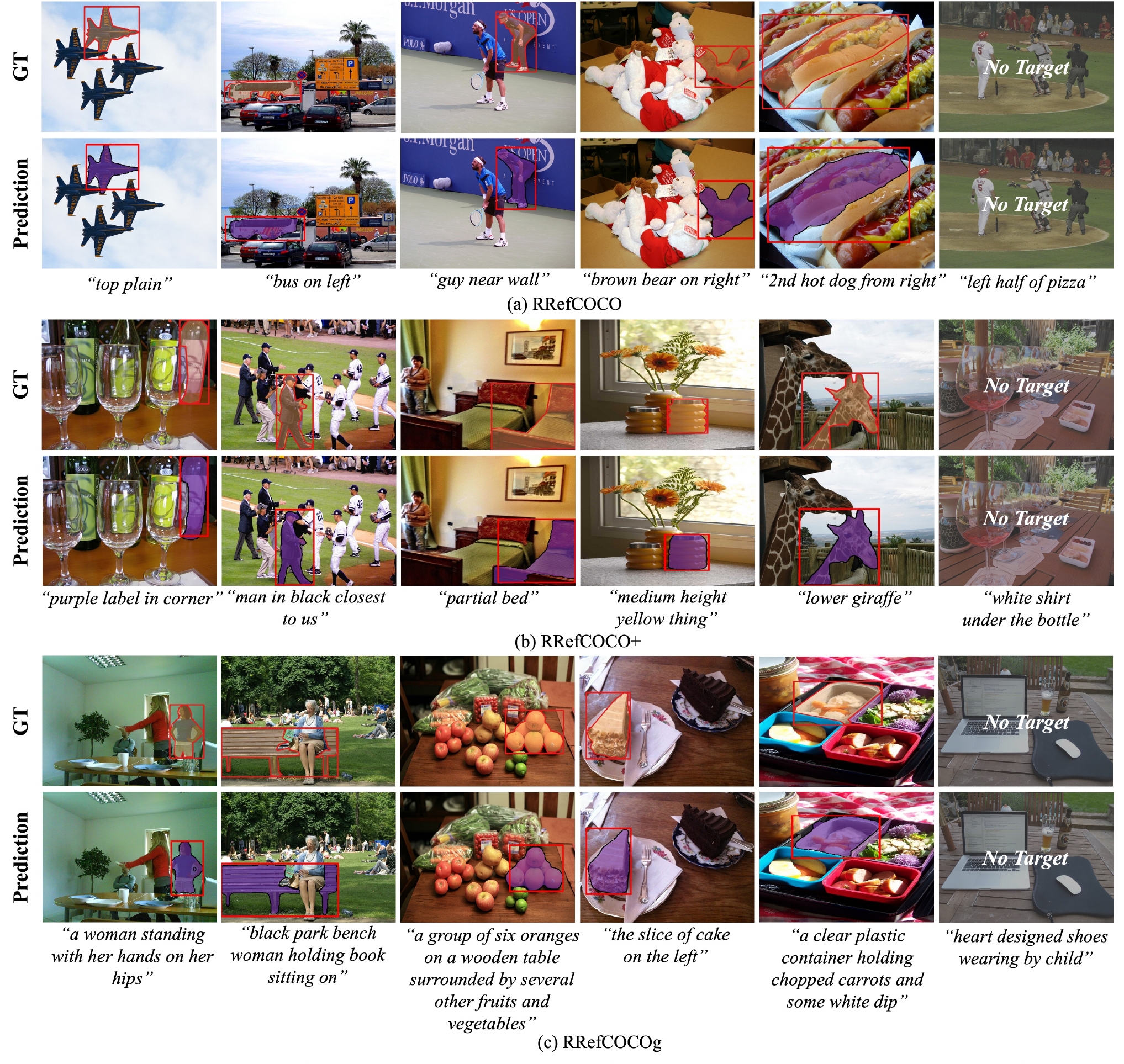}
\caption{Visualization on RRefCOCO/+/g \cite{wu2024toward} dataset.}
\label{fig_7}
\end{figure*}

\textbf{Visualization on GVG task.}
In Figure \ref{fig_5}, we further present qualitative results on gRefCOCO \cite{liu2023gres}, where SSDE performs detection and segmentation in a unified manner. For the same image, we apply expressions corresponding to single-target, multi-target, and non-target scenarios. The results indicate that SSDE maintains high accuracy across all settings, including challenging cases with missing targets or densely distributed similar instances. In Figure \ref{fig_7}, additional visualizations on RRefCOCO \cite{wu2024toward} further confirm its strong localization capability.

We also provide visualizations on the Ref-ZOM \cite{hu2023beyond} dataset, which contains complex multi-instance scenes, as shown in Figure \ref{fig_6}. SSDE demonstrates robust understanding of referring expressions, particularly in scenarios involving more than three targets, where it can still accurately localize all relevant instances. This performance underscores that query representations enriched with fine-grained visual attributes enhance inter-instance semantic discrimination, while instance-level structural constraints ensure spatial independence, effectively mitigating instance adhesion and spatial ambiguity.

\end{document}